\documentclass[runningheads]{llncs}

\usepackage{eccv}

\usepackage{eccvabbrv}
\usepackage{wrapfig}
\usepackage{graphicx}
\usepackage{booktabs}
\usepackage[utf8]{inputenc}
\usepackage{booktabs} 
\usepackage{multirow} 
\usepackage[table]{xcolor} 
\usepackage{adjustbox} 
\usepackage{amsmath} 
\usepackage[accsupp]{axessibility}  

\usepackage{hyperref}
\usepackage{orcidlink}

\begin{document}

\title{MAVFusion: Efficient Infrared and Visible Video Fusion via Motion-Aware Sparse Interaction}
\titlerunning{MAVFusion: Efficient Infrared and Visible Video Fusion}

\author{
Xilai Li$^{*,1}$ \and
Weijun Jiang$^{*,1}$ \and
Xiaosong Li$^{\dagger,1}$ \and
Yang Liu$^{1}$ \and
Hongbin Wang$^{2}$ \and
Tao Ye$^{3}$ \and
Huafeng Li$^{2}$ \and
Haishu Tan$^{1}$
}

\authorrunning{X. Li et al.}

\institute{
Foshan University, Foshan, China\\
\email{20210300236@stu.fosu.edu.cn}, \email{lixiaosong@buaa.edu.cn}
\and
Kunming University of Science and Technology, Kunming, China
\and
China University of Mining and Technology, Beijing, China
}

\begingroup
\renewcommand\thefootnote{}\footnotetext{$^*$ Equal contribution.}
\footnotetext{$^\dagger$ Corresponding author.}
\endgroup

\maketitle

\begin{abstract}
Infrared and visible video fusion combines the object saliency from infrared images with the texture details from visible images to produce semantically rich fusion results. However, most existing methods are designed for static image fusion and cannot effectively handle frame-to-frame motion in videos. Current video fusion methods improve temporal consistency by introducing interactions across frames, but they often require high computational cost. To mitigate these challenges, we propose MAVFusion, an end-to-end video fusion framework featuring a motion-aware sparse interaction mechanism that enhances efficiency while maintaining superior fusion quality. Specifically, we leverage optical flow to identify dynamic regions in multi-modal sequences, adaptively allocating computationally intensive cross-modal attention to these sparse areas to capture salient transitions and facilitate inter-modal information exchange. For static background regions, a lightweight weak interaction module is employed to maintain structural and appearance integrity. By decoupling the processing of dynamic and static regions, MAVFusion simultaneously preserves temporal consistency and fine-grained details while significantly accelerating inference. Extensive experiments demonstrate that MAVFusion achieves state-of-the-art performance on multiple infrared and visible video benchmarks, achieving a speed of 14.16\,FPS at $640 \times 480$ resolution. The source code is available at \href{https://github.com/ixilai/MAVFusion}{https://github.com/ixilai/MAVFusion}.
\end{abstract}

\section{Introduction}
\label{sec:intro}

In visual perception tasks, a single modality often cannot fully represent all the information in a scene. Practical applications usually require combining complementary information from different modalities to overcome the limitations of any single modality \cite{r3,r131,r143,r146,r148}. As an important branch of multi-modal visual fusion, infrared and visible fusion has significant practical value \cite{r112,r115,r116,r12}.

Although significant progress has been made in infrared and visible fusion, most existing studies still focus on image-level tasks, such as cross-modal interaction \cite{r24,r32,r144}, joint optimization with downstream tasks \cite{r114,r17,r121}, and degraded-scene fusion \cite{r91,r65,r111}. These methods usually ignore temporal dependencies across frames and lack awareness of motion and dynamic changes. As a result, they may produce temporal jitter and fail to fully exploit motion cues in video scenarios. Compared with single-frame fusion, multi-modal video fusion can better capture object motion and temporal variations by leveraging inter-frame correlations. This improves the coherence of fusion results and provides useful temporal references for cross-modal interaction in dynamic scenes.

\begin{figure}[t]
  \centering
   \includegraphics[width=1\linewidth]{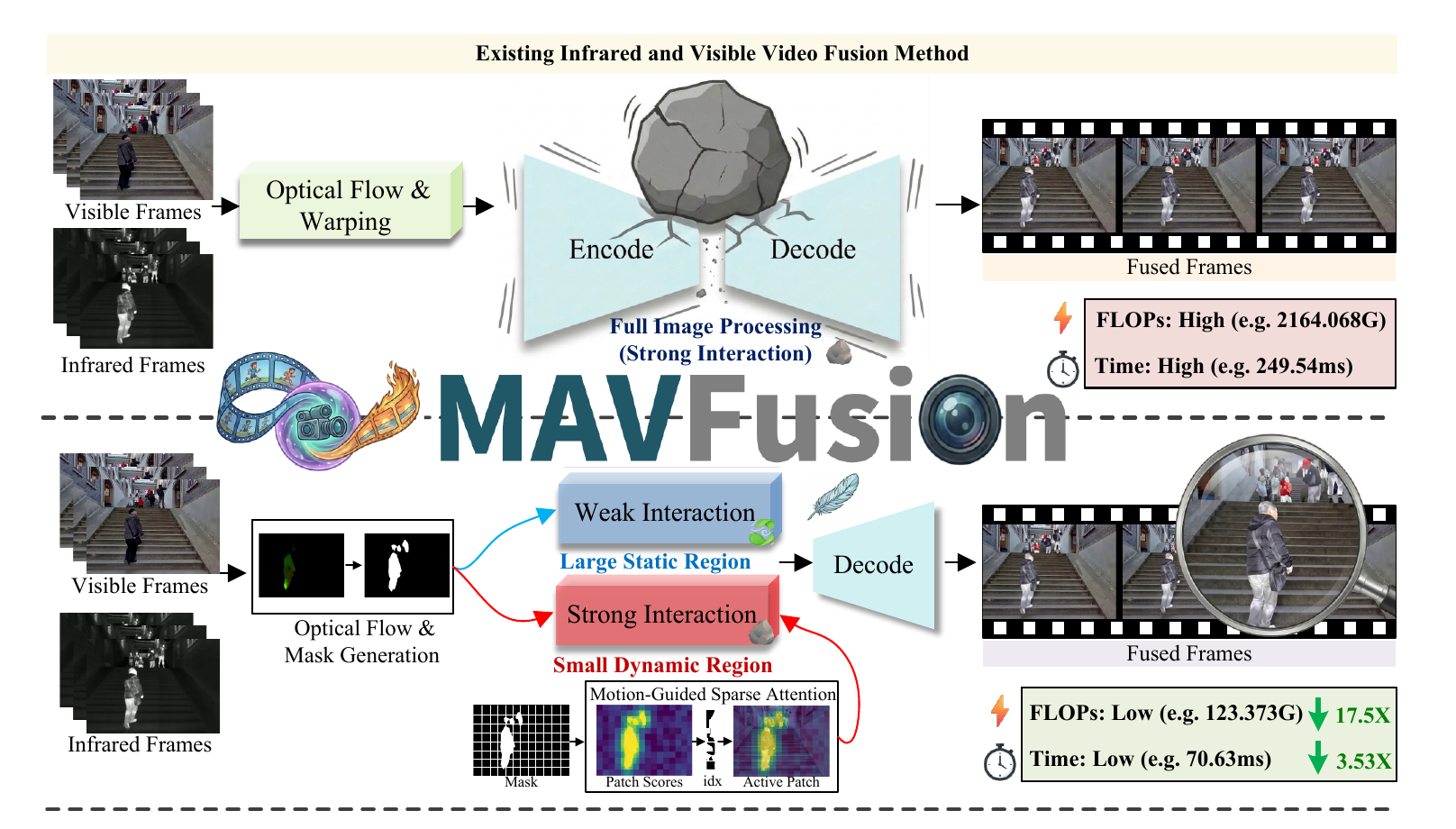}
   \caption{Comparison of the proposed algorithm with existing video fusion method \cite{r122} in terms of multi-modal interaction strategies and computational efficiency.}
   \label{fig1}
\end{figure}

Most existing video fusion methods \cite{r122,r123} rely on global modeling to maintain temporal consistency. This approach requires spatio-temporal interaction over large areas. Although it captures dynamic relationships, it also leads to high computational cost. In addition, dense attention mechanisms usually treat all regions equally.  As shown in \cref{fig1}, this is inefficient for long sequences, because much computation is spent on less important background areas instead of focusing on key moving objects that truly need strong fusion.
Furthermore, current methods do not fully use an important advantage of video: motion information can directly indicate where fusion should focus. Motion cues, such as optical flow, can reveal dynamic changes in the scene and help locate moving regions that contain important objects, such as pedestrians and vehicles. In infrared and visible fusion, strong cross-modal interaction and precise alignment are often most needed in these regions. In contrast, static background areas usually occupy most of the scene, and the modality with richer details (often the visible image) naturally provides better structure and texture. Applying excessive interaction in these static areas, or forcing information from a weaker or noisier modality, may disturb the original structure and semantics, leading to artifacts or unstable results.
These observations suggest that video fusion requires a new framework that allocates computation according to motion cues and balances strong interaction in key dynamic regions with stable preservation of static backgrounds.

To address the above issues, we propose an efficient infrared and visible video fusion method based on motion-aware sparse attention interaction (MAVFusion). Our method focuses on two aspects: improving temporal stability through motion-based alignment, and reducing computational cost by separating the interaction of static and dynamic regions.
First, we design a lightweight Motion-Aware Feature Alignment Module (MAFM) to improve cross-frame consistency and reduce ghosting. We use a frozen SEA-RAFT model \cite{r133} to estimate optical flow from visible frames for coarse alignment. Then, we predict residual optical flow in the feature space to refine the alignment and reduce errors between infrared and visible images. A motion mask is generated from the optical flow magnitude. The temporally aggregated features are injected through a motion-mask soft gate, so that motion regions benefit from multi-frame cues while static regions mainly rely on current-frame features to avoid unnecessary cross-frame mixing.
Second, we propose a static–dynamic decoupled multi-modal fusion module. The fusion process is divided into a weak interaction branch for static regions and a strong interaction branch for dynamic regions. In the static branch, convolutional modules are used for local modeling, which preserves background structure and texture details in a stable and low-cost way, ensuring natural and consistent overall appearance. In the dynamic branch, guided by the motion mask, sparse attention is applied only to selected Top-K patches in key regions.
Our main contributions are as follows:

\begin{itemize}
\item We propose an infrared and visible video fusion framework based on motion-aware sparse attention. The framework separates static and dynamic regions to allocate computation more efficiently, achieving good fusion quality with fast inference.

\item We propose a lightweight motion-aware feature alignment module that enhances cross-frame consistency in motion regions while avoiding noise in static areas, effectively reducing ghosting in multi-frame fusion.

\item We design a motion–guided sparse attention module that restricts complex interactions to important motion regions. This allows dynamic targets to build long-range connections with the global context, improving cross-modal fusion while reducing unnecessary computation.

\item Extensive experiments on multiple benchmarks consistently show that our method outperforms state-of-the-art approaches in fusion quality, temporal stability, and computational efficiency, striking an excellent balance between performance and speed.
\end{itemize}

\section{Related Work}

\subsection{Infrared and Visible Image Fusion}
Existing infrared and visible image fusion (IVIF) methods mainly include autoencoder based and generative model-based approaches. Autoencoder-based methods \cite{r77,r124,r125,r78} extract key features from source images and perform fusion using deep models, while generative model-based methods \cite{r4,r127,r128} learn latent image distributions to produce high-quality fusion results. Recent research further extends IVIF to real-world degradations, including noise \cite{r92}, motion blur \cite{r129}, and adverse weather \cite{r91,r130}. Some methods \cite{r117,r104} also exploit semantic information and task-driven modeling to improve fusion performance. However, these approaches still focus on single images and do not capture temporal information, limiting their effectiveness for video fusion.

\subsection{Infrared and Visible Video Fusion}

Compared to the well-studied field of static image fusion, infrared and visible video fusion must not only integrate complementary information within each frame but also maintain motion consistency and temporal stability across frames \cite{r132,r122,r123,r134}. For example, Guo et al. \cite{r132} reduced the effect of intra-frame feature discrepancies on fusion consistency using a hierarchical fusion strategy. Zhao et al. \cite{r122} combined multi-frame learning with optical flow feature registration in a single framework, producing fusion results with better temporal coherence.
Although these methods advance video fusion, they rely on global spatio-temporal modeling or dense attention mechanisms to ensure temporal consistency. This approach significantly increases computational cost and wastes processing power on unimportant background regions in long sequences.

\section{Methodology}

\begin{figure}[t]
  \centering
   \includegraphics[width=1\linewidth]{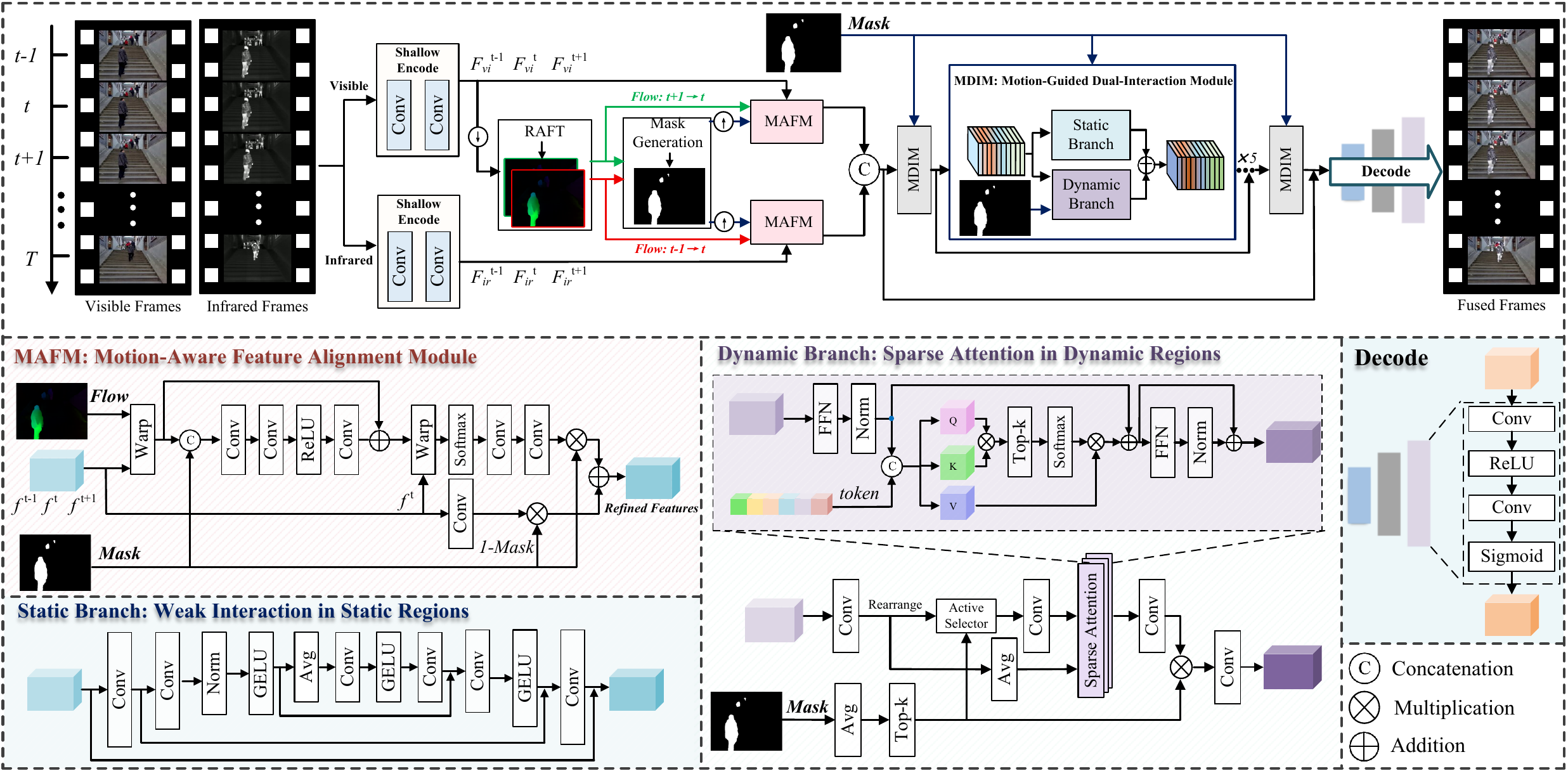}
   \caption{The overall framework of the proposed algorithm.}
   \label{fig2}
\end{figure}

The overall structure of the proposed algorithm is shown in \cref{fig2}. This framework consists of four core stages: shallow feature extraction and motion estimation, Motion-Aware Feature Alignment Module (MAFM), Motion-Guided Dual-Interaction Module (MDIM), and image reconstruction decoder.

\subsection{Motion-Aware Feature Alignment Module}

In dynamic video fusion tasks, cross-modal spatial misalignment and object motion often cause ghosting and blur in the fusion results. To address this problem, we propose a lightweight MAFM module. This module improves alignment through cross-modal guidance and a coarse-to-fine warping strategy. The alignment process includes two stages to gradually reduce spatial differences between adjacent frames.

\textbf{Multi-frame Coarse Alignment:}
First, to mitigate the prohibitive computational cost and large motion amplitudes at high resolutions \cite{r119,r120}, we estimate optical flow via SEA-RAFT \cite{r133} at a lower resolution and upsample it for initial spatial alignment. The coarse feature alignment is defined as:
\begin{equation}
\tilde{f}_i = \mathcal{W}(f_i, \phi_i), \quad i \in \{t-1, t, t+1\}
\end{equation}
where $\mathcal{W}(\cdot)$ denotes the bilinear interpolation warping operator, and $f_i$ represents the feature at time $i$. For adjacent frames, $\phi_{\text{prev}}$ and $\phi_{\text{next}}$ denote the optical flows from frame $t-1$ to $t$ and from frame $t+1$ to $t$, respectively. For the current frame, we use their average $\phi_t=(\phi_{\text{prev}}+\phi_{\text{next}})/2$ as an initial motion estimate.

\textbf{Cross-Modal Residual Refinement:}
To further reduce flow errors, we introduce features from the other modality as spatial anchors for cross-modal guidance. The module concatenates the anchor features, the three coarsely aligned frames, and the original flow field, and then predicts the residual flow $\Delta\phi$ using depthwise separable convolutions:
\begin{equation}
\Delta \phi = \mathcal{H}_{\text{refine}} 
\left(
\left[
f_{\text{anchor}}, 
\tilde{f}_{t-1}, 
\tilde{f}_{t}, 
\tilde{f}_{t+1}, 
\phi_{\text{prev}}, 
\phi_{\text{next}}
\right]
\right),
\end{equation}
where $\mathcal{H}_{\text{refine}}$ represents the refinement sequence involving channel compression and spatial refinement. Finally, the refined feature $\hat{f}_t$ is generated by applying the residual compensation:
\begin{equation}
\hat{f}_t = \mathcal{W}\left(f_t, \phi_t + \Delta \phi \right).
\end{equation}
After obtaining the aligned feature sequence, we use an efficient temporal aggregation strategy instead of complex 3D convolutions. By learning temporal importance weights $\omega_i$, which are initialized to one, Softmax-normalized along the temporal dimension in the forward pass, and optimized end-to-end with the overall training objective, MAFM adaptively estimates the contribution of each aligned frame:
\begin{equation}
F_{\text{agg}} = 
\sum_{i \in \{t-1, t, t+1\}}
\operatorname{Softmax}(\boldsymbol{\omega})_i\,\hat{f}_i .
\end{equation}
Here, $\hat{f}_i$ denotes the aligned feature used for temporal aggregation, where the center frame uses the refined feature $\hat{f}_t$, while the adjacent frames retain the coarsely aligned features. Finally, to maintain background stability while exploiting temporal information, MAFM does not directly apply the temporally aggregated features to all regions. Instead, the motion mask is used as a soft gate to inject temporal aggregation into motion regions, while static regions mainly rely on the current-frame features. This motion-gated fusion suppresses unnecessary cross-frame mixing and preserves stable background structures, while allowing dynamic regions to benefit from temporal cues.

\subsection{Motion-Guided Dual-Interaction Module}

\begin{figure}[t]
  \centering
   \includegraphics[width=0.9\linewidth]{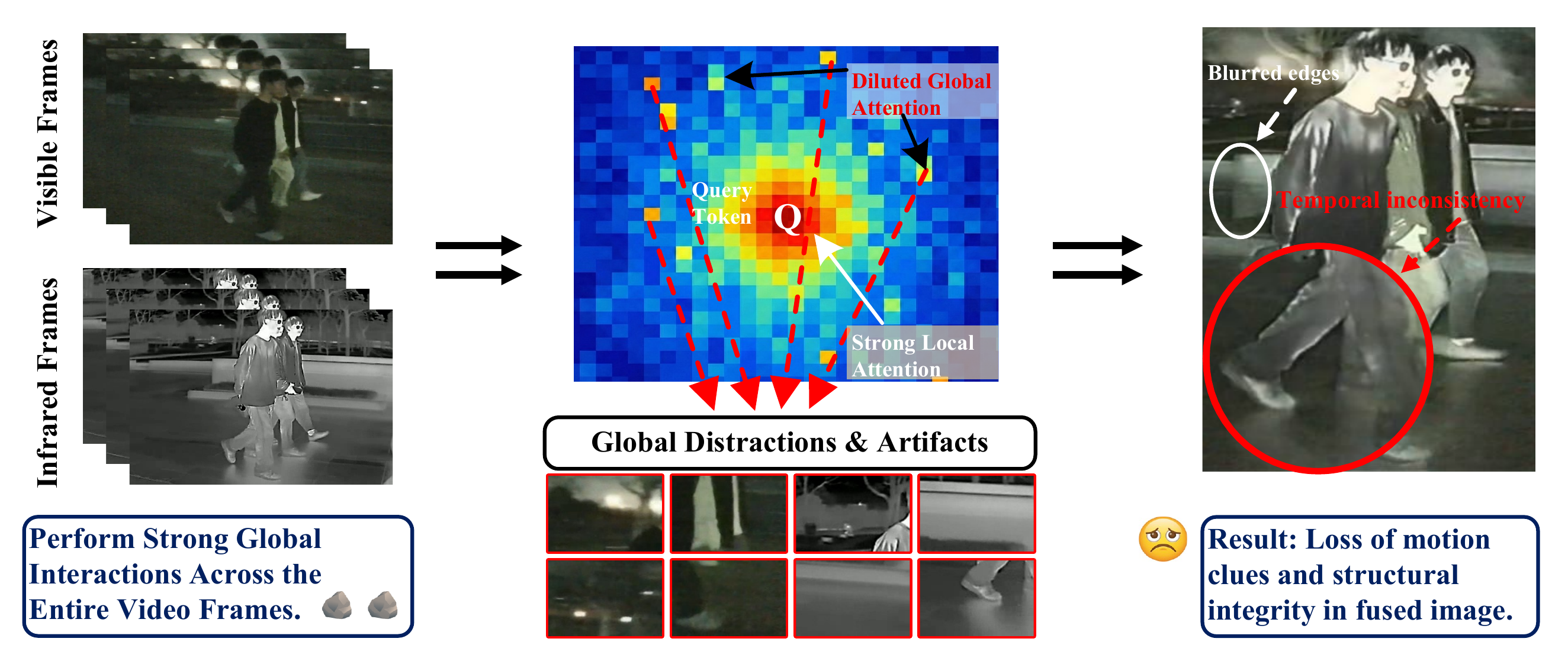}
   \caption{Effect of Global Strong Interaction on Motion Cues and Structural Integrity.}
   \label{fig_method}
\end{figure}

In infrared and visible fusion tasks, scenes often show clear static-dynamic differences: background regions are usually static with mostly redundant texture, while motion regions contain salient cross-modal features and important semantic targets. Existing video fusion methods apply uniform attention to all features, which leads to quadratic computational overhead. Furthermore, as shown in \cref{fig_method}, in long-sequence, high-resolution scenarios, attention must distribute weights across many tokens, causing salient and background regions to compete and resulting in diffuse attention. This can weaken local structures and reduce temporal consistency.
To address this, we propose a Motion-Guided Dual-Interaction Module. It applies weak interaction to static regions and sparse strong interaction to dynamic regions, enabling content-adaptive allocation of computation and more effective feature reconstruction.

\subsubsection{Dynamic Branch: Sparse Attention in Dynamic Regions}

Dynamic regions usually contain the most critical moving targets and salient semantic information in a scene, which require modeling long-range cross-modal dependencies. 
We employ an optical-flow-based motion mask 
$M \in \mathbb{R}^{H \times W}$ 
as a physical prior to guide the network to perform strong interactions only within active dynamic regions.

First, the input feature map is partitioned into non-overlapping $p \times p$ patches, resulting in a total of $N$ patches. 
By applying adaptive average pooling with the same spatial scale to the mask $M$, we obtain a patch-level saliency score vector $S \in \mathbb{R}^{N}$. 
Based on a predefined retention ratio $\tau=0.3$ (Top-K ratio), we set $k=\lfloor N\tau\rfloor$ and select the top-$k$ active patches, forming an index set 
$\Omega \subset \{1,\dots,N\}$, while retaining their corresponding saliency scores as soft weights $W$:
\begin{equation}
S = \text{AvgPool}_{p \times p}(M), 
\quad 
\Omega, W = \text{TopK}(S, \lfloor N \cdot \tau \rfloor).
\end{equation}
It is worth noting that this relative ranking strategy is especially important for handling ego-motion or camera shake. Instead of a hard threshold, the Top-$K$ selection adaptively isolates foreground objects with the strongest relative motion, keeping computation limited and preventing costly full-frame processing.

Since the identification and enhancement of dynamic targets often require reference to global contextual cues, 
we spatially average each patch feature to form a set of pooled global context tokens 
$T_{\text{global}}=\{T_i\}_{i=1}^{N}$:
\begin{equation}
T_i = \operatorname{AvgPool}\left(X_{\text{patch}}^{(i)}\right), 
\quad i=1,\dots,N.
\end{equation}
Subsequently, we construct an asymmetric joint sparse attention mechanism. 
The query vectors are generated only from the activated salient patch features $X_{\Omega}$, 
while the keys and values are provided by the pooled global context tokens $T_{\text{global}}$. 
Since only $k$ active patches are used as queries and all $N$ pooled patch tokens serve as global keys and values, 
this design reduces the computational complexity from $\mathcal{O}(N^2)$ to $\mathcal{O}(k \cdot N)$, 
achieving an effective balance between efficiency and receptive field:
\begin{equation}
F_{\text{attn}} = 
\operatorname{Softmax}
\left(
\frac{Q(X_{\Omega})\, K(T_{\text{global}})^{\top}}
{\sqrt{d}}
\right)
V(T_{\text{global}}).
\end{equation}

\subsubsection{Static Branch: Weak Interaction in Static Regions}

For static background regions, complex global self-attention mechanisms are often unnecessary. 
In the static branch, we employ lightweight depthwise separable convolutions together with standard $3 \times 3$ convolutions. 
This design constrains the receptive field to local neighborhoods for weak interaction, 
aiming to efficiently extract and preserve low-level high-frequency textures while significantly reducing computational cost.

\subsubsection{Adaptive Reconstruction and Fusion of Dynamic and Static Features}

After obtaining the static texture features $F_{\text{static}}$ and the dynamic salient features $F_{\text{attn}}$, 
they must be seamlessly fused in the spatial domain to avoid hard boundary artifacts.
We first use the saliency soft weights $W$ to reweight the dynamic features and then apply an index-copy operation (denoted as $\mathcal{R}$) 
to scatter the sparse patches back to their original spatial locations. 
Finally, the interpolated mask $M$ serves as a spatial gating signal to achieve adaptive reconstruction and smoothing:
\begin{equation}
Y = F_{\text{static}} + \text{Smooth}\big( M \odot \mathcal{R}(F_{\text{attn}} \odot W) \big).
\end{equation}
Here, $\text{Smooth}(\cdot)$ denotes a $3 \times 3$ smoothing convolution layer, 
which alleviates stitching artifacts caused by multi-scale patch partitioning 
and ensures a natural spatial and semantic transition in the final fused feature $Y$.

\subsection{Loss Function}

To generate high-quality fused video sequences with spatial fidelity and temporal consistency, we define the total loss $\mathcal{L}_{\text{total}}$ as a weighted combination of the spatial loss $\mathcal{L}_{\text{spatial}}$ and the temporal consistency loss $\mathcal{L}_{\text{temp}}$:
\begin{equation}
    \mathcal{L}_{\text{total}} = \mathcal{L}_{\text{spatial}} + \gamma \mathcal{L}_{\text{temp}},
\end{equation}
where $\gamma$ is a hyper-parameter used to balance spatial reconstruction and temporal stability in video sequences.

\subsubsection{Spatial Fidelity Constraint}
The spatial loss $\mathcal{L}_{\text{spatial}}$ primarily serves to integrate complementary and salient information from the two modalities. Specifically, $\mathcal{L}_{\text{spatial}}$ consists of a pixel similarity loss \cite{r122} and a structural similarity loss \cite{r135}, preserving infrared saliency and visible details in the fused frame $I^F$.

\subsubsection{Temporal Consistency Constraint}
To mitigate inter-frame flickering and ensure smooth visual transitions in the output video, we introduce a temporal consistency loss $\mathcal{L}_{\text{temp}}$ \cite{r122}. This term explicitly enforces frame-to-frame stability by penalizing misalignments between adjacent fused frames using estimated optical flow fields $\mathcal{O}$. To avoid unreliable gradients in occluded or poorly aligned regions, we incorporate validity masks $M_{prev}^t$ and $M_{next}^t$ to identify reliable pixels \cite{r122}. The loss is defined as:
\begin{equation}
\begin{aligned}
\mathcal{L}_{\text{temp}} 
=&\; \mathbb{E}_{p \in M_{prev}^t} 
\left[ \left| I_t^F(p) - \mathcal{W}(I_{t-1}^F, \mathcal{O}_{t-1 \rightarrow t}^F)(p) \right|_1 \right] \\
&+ \mathbb{E}_{p \in M_{next}^t} 
\left[ \left| I_t^F(p) - \mathcal{W}(I_{t+1}^F, \mathcal{O}_{t+1 \rightarrow t}^F)(p) \right|_1 \right].
\end{aligned}
\end{equation}
where $p$ denotes the pixel coordinates, and $\mathcal{W}(\cdot, \mathcal{O})$ denotes the backward warping operation guided by the flow field $\mathcal{O}$.

\section{Experiments}

\subsection{Experiment Settings}

\subsubsection{Datasets and Metrics}

To evaluate the performance of the proposed method on infrared and visible video fusion, we conduct experiments on three public video datasets: M3SVD \cite{r123}, HDO \cite{r136}, and VTMOT \cite{r137}. The M3SVD dataset contains 30 video sequences, HDO contains 24 sequences, and VTMOT contains 90 sequences. 
For each dataset, we randomly select five video sequences for testing, and use the remaining sequences for training.
Furthermore, to comprehensively evaluate the fusion performance, we adopt eight objective metrics, including image quality metrics and a video smoothness metric. 
The image quality metrics include Gradient-Based Fusion Performance ($Q_G$), Image Fusion Metric Based on a Multiscale Scheme ($Q_M$), Image Fusion Metric Based on Phase Congruency ($Q_P$), Piella’s Metric ($Q_S$), Chen--Blum Metric ($Q_{CB}$), Visual Information Fidelity ($VIF$), and Normalized Weighted Performance Metric ($Q^{AB/F}$) \cite{r69}. 
For video smoothness evaluation, we use Motion Smoothness with Dual Reference Videos (MS2R) \cite{r122}.

\subsubsection{Comparative Methods}

To validate the effectiveness of the proposed method, we compare it with seven state-of-the-art image fusion methods and two video fusion methods.
The image fusion methods include UP-Fusion \cite{r117}, TDFusion \cite{r138}, SAGE \cite{r139}, GIFNet \cite{r114}, RFFusion \cite{r140}, UMCFuse \cite{r116}, and FreeFusion \cite{r115}. The video fusion methods are VideoFusion \cite{r123} and UniVF \cite{r122}. 
All methods are evaluated under the same experimental settings to ensure a fair comparison. 

\subsubsection{Training Details}

All experiments are conducted on a machine equipped with four NVIDIA GeForce RTX 3090 GPUs. During training, input frames are randomly cropped to a size of $384 \times 384$. The batch size is set to 32 with gradient accumulation enabled.
We use the Adam optimizer with an initial learning rate of $1.0\times10^{-4}$. After 40,000 iterations, the learning rate is exponentially decayed to 1\% of its initial value.

\begin{figure}[t]
  \centering
   \includegraphics[width=1\linewidth]{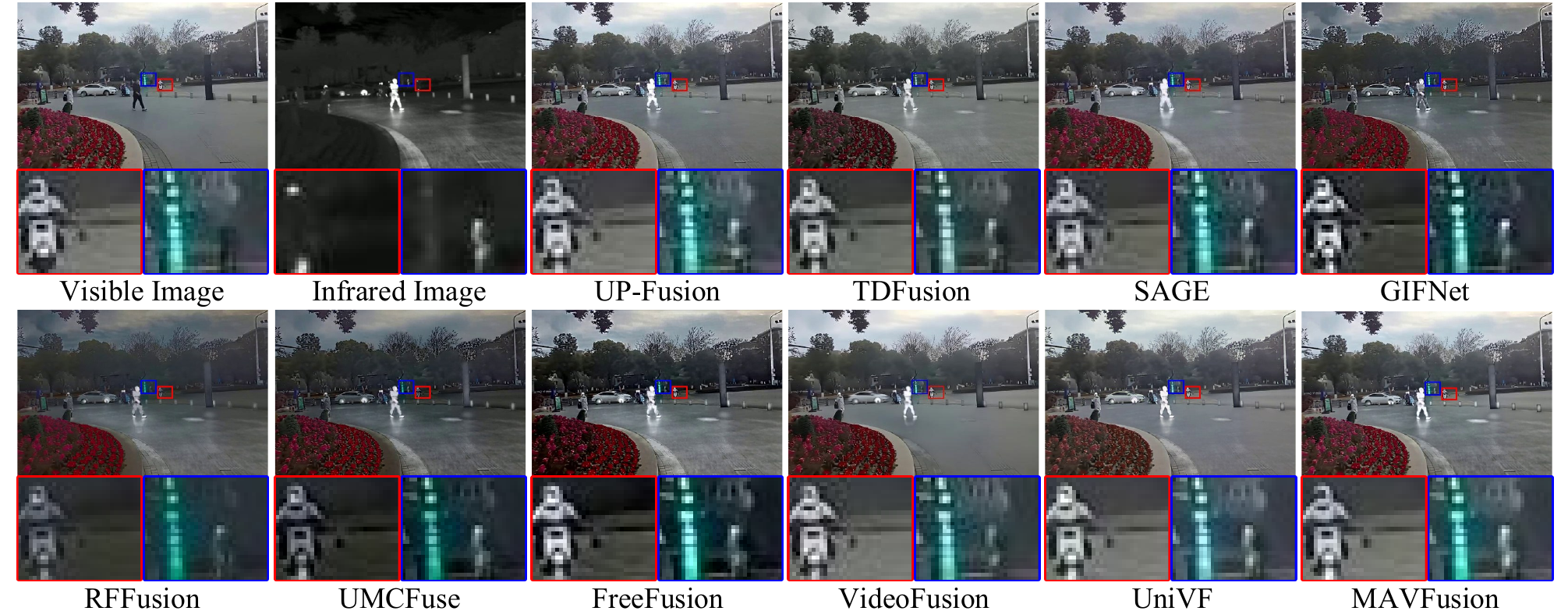}
   \caption{Qualitative comparison of all methods on the M3SVD dataset.}
   \label{fig3}
\end{figure}
\begin{figure}[t]
  \centering
   \includegraphics[width=1\linewidth]{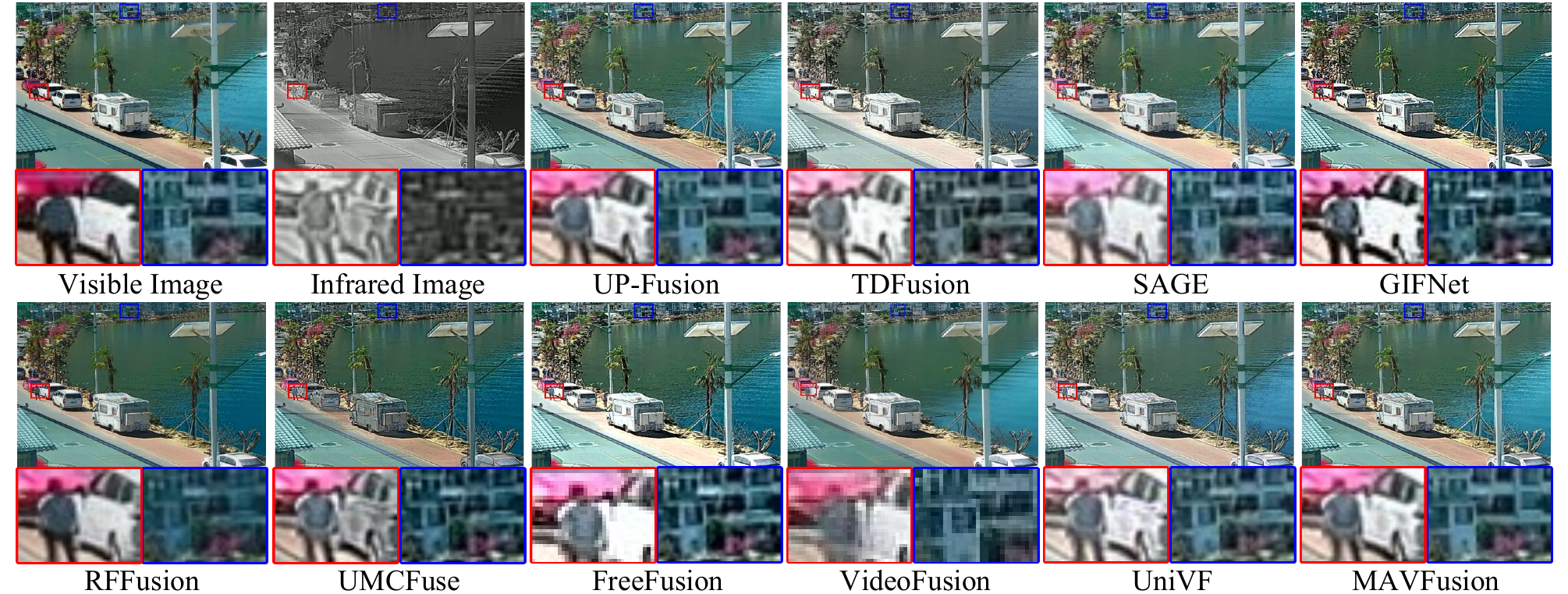}
   \caption{Qualitative comparison of all methods on the HDO dataset.}
   \label{fig4}
\end{figure}
\subsection{Qualitative Comparative Experiment}

\begin{figure}[t]
  \centering
   \includegraphics[width=1\linewidth]{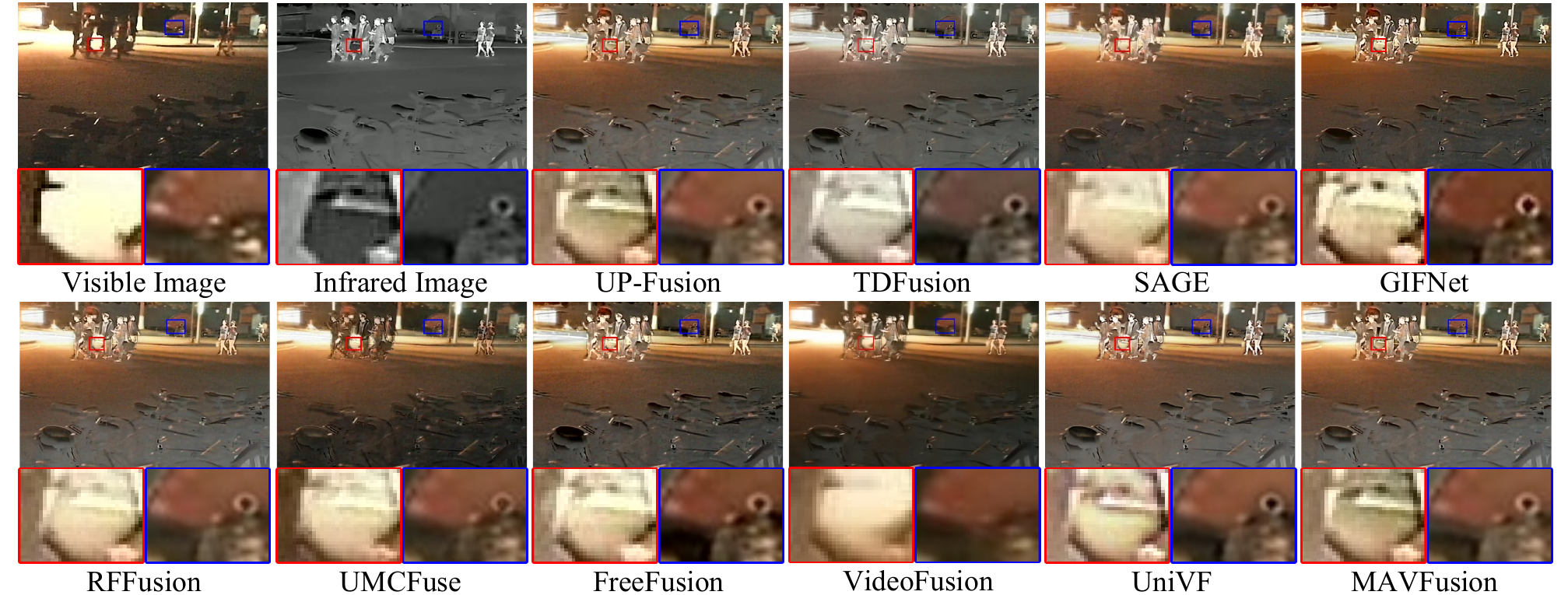}
   \caption{Qualitative comparison of all methods on the VTMOT dataset.}
   \label{fig5}
\end{figure}

\cref{fig3,fig4,fig5} show a qualitative comparison of the proposed MAVFusion algorithm with competing methods on three video fusion datasets (M3SVD, HDO, VTMOT). For each result, we provide zoomed-in views of both moving and static regions.
For dynamic objects, methods that ignore temporal continuity, such as UP-Fusion, SAGE, and GIFNet, often produce ghosting and blurred edges. RFFusion, UMCFuse, and FreeFusion alleviate ghosting by relying more on infrared background information, but this weakens spatial contrast and blurs scene structures.
In contrast, video fusion methods such as UniVF and MAVFusion better suppress ghosting and preserve contrast. However, UniVF may over-introduce smooth infrared information into static regions, weakening visible textures and overall scene expressiveness.
MAVFusion separates dynamic and static processing for region-adaptive fusion, effectively mitigating spatio-temporal conflicts in video fusion.

\subsection{Quantitative Comparative Experiment}

\cref{tab1} reports the quantitative comparison results across three datasets, with the top two performances highlighted. The proposed algorithm achieves state-of-the-art scores across most image quality metrics. Several competitive methods, such as UP-Fusion, TDFusion, and UniVF, also show strong performance in image gradients, structural integrity, and perceptual quality, demonstrating their ability to preserve spatial details. Additionally, our weak-interaction design for background regions proves effective, indicating that sparse-textured backgrounds require only minimal interaction to maintain high-fidelity fusion. The strong MS2R result further demonstrates good temporal continuity. Sparse strong interaction in dynamic regions helps capture motion-aware multimodal cues, reduce ghosting, and enhance spatial contrast.

\definecolor{bestcolor}{RGB}{255, 230, 230} 
\definecolor{secondcolor}{RGB}{230, 255, 230} 

\begin{table*}[t]
\centering
\footnotesize 
\setlength{\tabcolsep}{3.8pt}
\caption{Quantitative comparison on M3SVD, HDO, and VTMOT datasets. \textbf{Bold} and \colorbox{bestcolor}{red} indicate the best performance, while \colorbox{secondcolor}{green} indicates the second best.}
\label{tab1}

\begin{adjustbox}{max width=\textwidth}
\begin{tabular}{c | l | c | cccccccc}
\toprule\toprule
Type & Methods & Pub. & $Q_G$↑ & $Q_M$↑ & $Q_P$↑ & $Q_S$↑ & $Q_{CB}$↑ & $VIF$↑ & $Q^{AB/F}$↑ & $MS2R$↓ \\
\midrule
\multicolumn{11}{c}{\textbf{M3SVD Dataset}} \\
\midrule
\multirow{7}{*}{Image} 
 & UP-Fusion & \textit{AAAI 26} & 0.6153 & \cellcolor{secondcolor}0.8126 & 0.5391 & 0.8394 & \cellcolor{secondcolor}0.5173 & \cellcolor{secondcolor}0.3492 & 0.6935 & 0.1110 \\
 & TDFusion & \textit{CVPR 25} & \cellcolor{secondcolor}0.6703 & 0.5606 & 0.5346 & \cellcolor{secondcolor}0.8520 & 0.5163 & 0.3382 & 0.6962 & \cellcolor{secondcolor}0.1059 \\
 & SAGE & \textit{CVPR 25} & 0.4739 & 0.3088 & 0.4433 & 0.8114 & 0.4697 & 0.2974 & 0.6110 & 0.1130 \\
 & GIFNet & \textit{CVPR 25} & 0.3988 & 0.1966 & 0.3318 & 0.7537 & 0.4653 & 0.1969 & 0.4823 & 0.1169 \\
 & RFFusion & \textit{NeurIPS 25} & 0.4371 & 0.2539 & 0.4106 & 0.7085 & 0.4071 & 0.3313 & 0.3589 & 0.1107 \\
 & UMCFuse & \textit{TIP 25} & 0.6256 & 0.5049 & 0.4287 & 0.7818 & 0.5015 & 0.2807 & 0.5994 & 0.1127 \\
 & FreeFusion & \textit{TPAMI 25} & 0.4485 & 0.2338 & 0.4027 & 0.6702 & 0.4186 & 0.2009 & 0.5438 & 0.1215 \\
\midrule
\multirow{3}{*}{Video} 
 & VideoFusion & \textit{CVPR 26} & 0.3881 & 0.2832 & 0.4072 & 0.7964 & 0.4380 & 0.2631 & 0.5491 & 0.1593 \\
 & UniVF & \textit{NeurIPS 25} & 0.6376 & 0.5661 & \cellcolor{secondcolor}0.5529 & 0.8449 & 0.5169 & 0.3425 & \cellcolor{secondcolor}0.7049 & 0.1064 \\
 & \textbf{MAVFusion} & -- & \cellcolor{bestcolor}\textbf{0.6897} & \cellcolor{bestcolor}\textbf{1.1544} & \cellcolor{bestcolor}\textbf{0.6122} & \cellcolor{bestcolor}\textbf{0.8549} & \cellcolor{bestcolor}\textbf{0.5535} & \cellcolor{bestcolor}\textbf{0.3814} & \cellcolor{bestcolor}\textbf{0.7550} & \cellcolor{bestcolor}\textbf{0.1058} \\
\midrule\midrule

\multicolumn{11}{c}{\textbf{HDO Dataset}} \\
\midrule
\multirow{7}{*}{Image} 
 & UP-Fusion & \textit{AAAI 26} & 0.5997 & 0.7588 & \cellcolor{secondcolor}0.5096 & \cellcolor{secondcolor}0.8441 & \cellcolor{secondcolor}0.5345 & \cellcolor{secondcolor}0.3186 & 0.6200 & \cellcolor{bestcolor}\textbf{0.9657} \\
 & TDFusion & \textit{CVPR 25} & \cellcolor{secondcolor}0.6153 & 0.6787 & 0.4659 & 0.8417 & 0.5155 & 0.2981 & 0.6081 & 0.9952 \\
 & SAGE & \textit{CVPR 25} & 0.3928 & 0.3536 & 0.3643 & 0.7528 & 0.5044 & 0.2674 & 0.4620 & 1.0622 \\
 & GIFNet & \textit{CVPR 25} & 0.3465 & 0.2565 & 0.2987 & 0.6959 & 0.5105 & 0.1909 & 0.3823 & 1.0546 \\
 & RFFusion & \textit{NeurIPS 25} & 0.4478 & 0.3637 & 0.3316 & 0.6773 & 0.4645 & 0.2619 & 0.4191 & 1.0267 \\
 & UMCFuse & \textit{TIP 25} & 0.5879 & 0.6375 & 0.4099 & 0.8185 & 0.5140 & 0.2795 & 0.5471 & 1.0110 \\
 & FreeFusion & \textit{TPAMI 25} & 0.4839 & 0.4183 & 0.3667 & 0.6797 & 0.4860 & 0.2124 & 0.5044 & 1.0020 \\
\midrule
\multirow{3}{*}{Video} 
 & VideoFusion & \textit{CVPR 26} & 0.3649 & 0.3372 & 0.3642 & 0.7391 & 0.4781 & 0.2616 & 0.4128 & 1.2651 \\
 & UniVF & \textit{NeurIPS 25} & 0.6125 & \cellcolor{secondcolor}0.7655 & 0.4953 & 0.8433 & 0.5300 & 0.3109 & \cellcolor{secondcolor}0.6364 & 1.0086 \\
 & \textbf{MAVFusion} & -- & \cellcolor{bestcolor}\textbf{0.6629} & \cellcolor{bestcolor}\textbf{0.9996} & \cellcolor{bestcolor}\textbf{0.5660} & \cellcolor{bestcolor}\textbf{0.8464} & \cellcolor{bestcolor}\textbf{0.5578} & \cellcolor{bestcolor}\textbf{0.3359} & \cellcolor{bestcolor}\textbf{0.6837} & \cellcolor{secondcolor}0.9826 \\
\midrule\midrule

\multicolumn{11}{c}{\textbf{VTMOT Dataset}} \\
\midrule
\multirow{7}{*}{Image} 
 & UP-Fusion & \textit{AAAI 26} & 0.5154 & 0.6226 & 0.3894 & 0.8486 & 0.3635 & 0.3798 & 0.6307 & 1.5948 \\
 & TDFusion & \textit{CVPR 25} & 0.5815 & 0.8928 & 0.4197 & \cellcolor{bestcolor}\textbf{0.8517} & 0.3899 & \cellcolor{bestcolor}\textbf{0.4147} & \cellcolor{secondcolor}0.6713 & \cellcolor{secondcolor}0.8346 \\
 & SAGE & \textit{CVPR 25} & 0.3285 & 0.3129 & 0.2168 & 0.7592 & 0.3819 & 0.2877 & 0.4169 & 0.9498 \\
 & GIFNet & \textit{CVPR 25} & 0.2963 & 0.2420 & 0.2223 & 0.7113 & \cellcolor{bestcolor}\textbf{0.4389} & 0.2115 & 0.3834 & 0.9402 \\
 & RFFusion & \textit{NeurIPS 25} & 0.4899 & 0.3836 & 0.2731 & 0.6741 & 0.4138 & 0.2715 & 0.4176 & 0.9082 \\
 & UMCFuse & \textit{TIP 25} & \cellcolor{secondcolor}0.5968 & \cellcolor{secondcolor}0.9231 & 0.3831 & 0.8272 & 0.4077 & 0.3534 & 0.6050 & 0.9939 \\
 & FreeFusion & \textit{TPAMI 25} & 0.5320 & 0.5073 & 0.3705 & 0.7881 & \cellcolor{secondcolor}0.4143 & 0.2950 & 0.5749 & 0.9775 \\
\midrule
\multirow{3}{*}{Video} 
 & VideoFusion & \textit{CVPR 26} & 0.2481 & 0.2784 & 0.1686 & 0.7448 & 0.3382 & 0.2832 & 0.3332 & 1.2241 \\
 & UniVF & \textit{NeurIPS 25} & 0.5724 & 0.8993 & \cellcolor{secondcolor}0.4276 & 0.8469 & 0.3892 & 0.3856 & 0.6705 & 0.8380 \\
 & \textbf{MAVFusion} & -- & \cellcolor{bestcolor}\textbf{0.6325} & \cellcolor{bestcolor}\textbf{0.9741} & \cellcolor{bestcolor}\textbf{0.4947} & \cellcolor{secondcolor}0.8504 & 0.4036 & \cellcolor{secondcolor}0.3885 & \cellcolor{bestcolor}\textbf{0.7182} & \cellcolor{bestcolor}\textbf{0.8301} \\
\bottomrule\bottomrule
\end{tabular}
\end{adjustbox}
\end{table*}

\begin{figure}[t]
  \centering
   \includegraphics[width=1\linewidth]{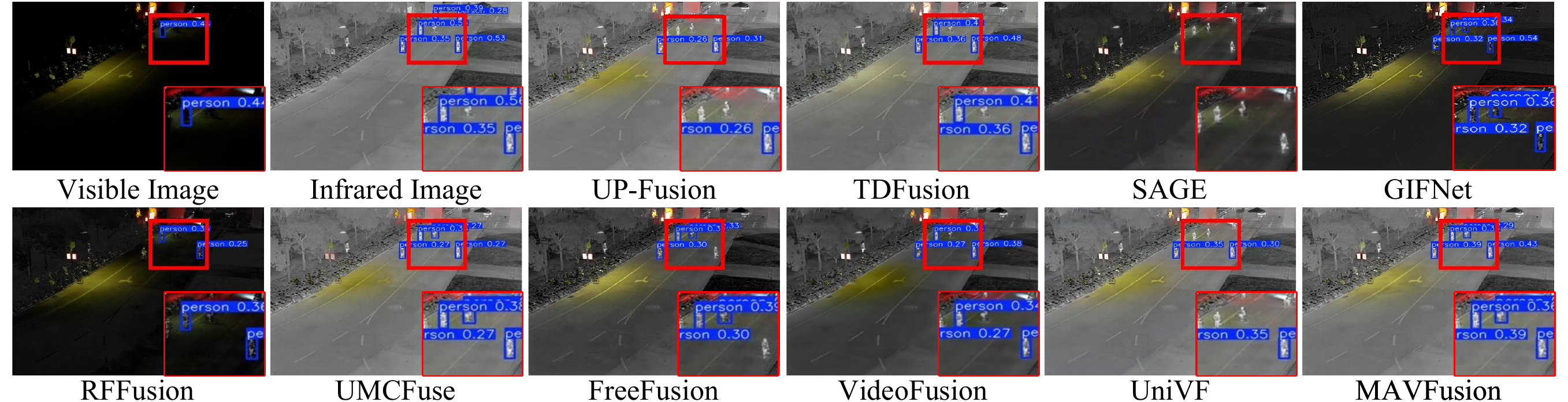}
   \caption{Detection accuracy comparison on fused results.}
   \label{fig6}
\end{figure}

\subsection{Downstream Task Experiments}

To evaluate the practical potential of the proposed algorithm, we conduct downstream object detection experiments using YOLO26 \cite{r142} as the base detector. This evaluation aims to compare the detection accuracy of various fused results across different scenarios.
As illustrated in \cref{fig6}, in low-illumination environments where visible information is nearly entirely lost, object detection becomes heavily reliant on the thermal saliency provided by the infrared modality. Competing methods such as SAGE, GIFNet, and RFFusion fail to maintain high target-to-background contrast due to the excessive integration of non-informative visible noise, which significantly diminishes target saliency.
Notably, our minimal intervention strategy for background regions suppresses interference from degraded visible inputs. By balancing modality integration and noise filtering, the method achieves the highest detection confidence across all targets, showing its advantage for high-level vision tasks.

\begin{table}[t]
\centering
\caption{Quantitative comparison of different ablation results.} 
\scriptsize
\label{tab2}
\begin{tabular}{l|cccccccc}
\toprule\toprule
Methods   & $Q_G$↑ & $Q_M$↑ & $Q_P$↑ & $Q_S$↑ & $Q_{CB}$↑ & $VIF$↑ & $Q^{AB/F}$↑ & $MS2R$↓ \\ \midrule
Full-DB   & 0.6276 & 0.9282 & 0.4897 & 0.8445 & 0.3943 & 0.3844 & 0.6991 & 0.9878 \\
Full-SB   & 0.6260 & 0.9550 & 0.4857 & \cellcolor{bestcolor}\textbf{0.8680} & 0.3996 & 0.3801 & 0.7108 & 0.9913 \\
w/ IM      & 0.5880 & 0.7863 & 0.4690 & 0.8413 & 0.4017 & 0.3825 & 0.6889 & 0.9842 \\
w/o SA    & 0.6080 & 0.8351 & 0.4861 & 0.8495 & 0.3984 & 0.3828 & 0.6975 & 0.9662 \\
w/o MAFM  & 0.5972 & 0.7820 & 0.4676 & 0.8501 & 0.3962 & 0.3806 & 0.6876 & 0.9799 \\
MAVFusion & \cellcolor{bestcolor}\textbf{0.6325} & \cellcolor{bestcolor}\textbf{0.9741} & \cellcolor{bestcolor}\textbf{0.4947} & 0.8504 & \cellcolor{bestcolor}\textbf{0.4036} & \cellcolor{bestcolor}\textbf{0.3885} & \cellcolor{bestcolor}\textbf{0.7182} & \cellcolor{bestcolor}\textbf{0.8301} \\ 
\bottomrule\bottomrule
\end{tabular}
\end{table}

\subsection{Ablation Studies}


\subsubsection{Analysis of the Region Separating Strategy}

To verify the necessity of separating dynamic and static regions, we design three variant models for comparison:

\begin{itemize}

\item \textbf{Full-DB (Full Dynamic Branch)}: 
We remove the mask guidance and input the whole image into the dynamic branch for strong interaction, to evaluate the effect of global strong interaction.

\item \textbf{Full-SB (Full Static Branch)}: 
We remove the dynamic branch and use only the static branch to process the whole image, to test the impact of lacking dynamic enhancement.

\item \textbf{w/ IM (Inverted Mask)}: 
We reverse the mask logic, applying strong interaction to static regions and weak processing to dynamic regions, to verify our spatial feature allocation.

\end{itemize}

\paragraph{Analysis of Full-DB.}

As shown in \cref{tab2}, applying strong interaction to the whole image (Full-DB) increases computation cost, but does not improve performance. Salient and background regions compete during attention computation, which scatters the attention distribution and weakens local structures. As a result, the scores of $Q_G$, $Q_M$, and $Q_S$ decrease.

\paragraph{Analysis of Full-SB.}

Removing the dynamic branch improves efficiency, but it ignores frame displacement in video sequences, as reflected by the clear deterioration of MS2R in \cref{tab2}. The model becomes a static weighting scheme, which causes overlap and conflicts around moving object boundaries. Although the $Q_S$ score increases due to more stable backgrounds, this comes at the cost of dynamic object saliency.

\paragraph{Analysis of w/ IM.}

In this variant, static background regions, which only need simple information complement, are processed with strong interaction. 
In this case, $Q_M$, $Q_P$, and $Q^{AB/F}$ drop significantly. 
This shows that the strategy not only wastes computation, but also lets smooth infrared background information interfere with clear visible textures, causing blur and loss of contrast.

\subsubsection{Analysis of the Sparse Attention}

We replace the original sparse attention with dense attention based on fully connected Softmax within each patch (w/o SA). As shown in \cref{tab2}, although the metric scores do not drop significantly, the results show degradation in details and contrast. This indicates that dense interaction covers more pixels, but processing redundant information does not effectively improve fusion performance and may smooth some local features.

\subsubsection{Analysis of the MAFM Module}

To verify the effectiveness of the MAFM module, we remove it and replace it with the multi-frame alignment strategy used in UniVF (w/o MAFM). As shown in \cref{tab2}, most quality metrics decrease, while MS2R becomes worse. Without anchor features as alignment references, the ability of the model to integrate cross-modal features is weakened. In addition, removing the dynamic-static separation inside the module allows background features to interfere with dynamic regions.

\subsection{Computational Efficiency Analysis}

To evaluate efficiency, we compare MAVFusion with representative methods in terms of FLOPs, parameters, and FPS. RFFusion, a diffusion-based method, is not included due to its fixed-size input requirement and high computational cost.

As shown in \cref{tab3}, MAVFusion is much more efficient than video-fusion baselines. At 640$\times$480, MAVFusion uses only 123.37G FLOPs, about 6.6\% of VideoFusion (1874.00G) and 5.7\% of UniVF (2164.07G), while achieving the highest FPS (14.16). At 1280$\times$720, MAVFusion requires only 267.88G FLOPs, about 4.8\% of VideoFusion and 3.8\% of UniVF, and runs 2.22$\times$ faster than VideoFusion and 4.56$\times$ faster than UniVF. Moreover, when the number of pixels increases by 3$\times$, the FLOPs of MAVFusion increase by only 2.17$\times$, showing better scalability than VideoFusion and UniVF. Although its speed is lower than very lightweight image-fusion models such as TDFusion and SAGE, this is reasonable because video fusion requires temporal modeling and optical-flow estimation. Overall, MAVFusion achieves a favorable balance between fusion performance and computational efficiency.

\begin{table}[t]
\centering
\caption{Comparison of computational complexity (FLOPs and parameters) and inference time (FPS) 
under different input resolutions.}
\renewcommand{\arraystretch}{1.2}
\setlength{\tabcolsep}{5pt}

\adjustbox{width=\textwidth}{
\begin{tabular}{c|c|c|ccc|ccc}
\hline
\hline
\multirow{2}{*}{Type} & \multirow{2}{*}{Methods} & \multirow{2}{*}{Pub} 
& \multicolumn{3}{c|}{640$\times$480} 
& \multicolumn{3}{c}{1280$\times$720} \\
\cline{4-9}
& & 
& FLOPs (G) & Params (M) & FPS 
& FLOPs (G) & Params (M) & FPS \\
\hline

\multirow{5}{*}{Image}
& UP-Fusion   & \textit{AAAI 26}   & 953.86 & 154.82 & 1.80  & 2830.84 & 154.82 & 0.62 \\
& TDFusion    & \textit{CVPR 25}   & 18.21  & 0.06   & 27.61 & 54.63   & 0.06   & 9.35 \\
& SAGE        & \textit{CVPR 25}   & 20.34  & 0.136  & 153.45& 61.00   & 0.136  & 51.63 \\
& GIFNet      & \textit{CVPR 25}   & 226.12 & 0.82   & 5.68  & 678.32  & 0.82   & 2.64 \\
& FreeFusion  & \textit{TPAMI 25}  & 451.65 & 5.67   & 12.81 & 1354.83 & 5.67   & 4.48 \\

\hline

\multirow{3}{*}{Video}
& VideoFusion & \textit{CVPR 26}    & 1874.00 & 6.73 & 7.83  & 5623.00 & 6.73 & 2.59 \\
& UniVF       & \textit{NeurIPS 25} & 2164.07 & 9.20 & 4.01  & 7055.34 & 9.20 & 1.26 \\

& \cellcolor{gray!15}MAVFusion 
& \cellcolor{gray!15}--- 
& \cellcolor{gray!15}123.37  
& \cellcolor{gray!15}9.90 
& \cellcolor{gray!15}14.16 
& \cellcolor{gray!15}267.88  
& \cellcolor{gray!15}9.90 
& \cellcolor{gray!15}5.74 \\

\hline
\hline
\end{tabular}
}
\label{tab3}
\end{table}

\subsection{Limitation}

Although MAVFusion shows clear advantages in efficiency and performance, it still has some limitations. First, severe sensor noise or extreme degradation may still impair the extraction of reliable modality-specific features, making it difficult for the model to completely suppress degradation artifacts in highly corrupted regions. Second, while the region separation strategy greatly reduces fusion network computation, the extra cost of the front-end optical flow estimator remains a challenge. We plan to address both issues in future work.

\section{Conclusion}

This paper proposes MAVFusion, an efficient infrared and visible video fusion method based on motion-aware sparse interaction. It achieves high-fidelity fusion and frame-to-frame interaction while maintaining efficiency. To address cross-modal spatial misalignment and ghosting or blur caused by object motion, we design a lightweight Motion-Aware Feature Alignment Module. In the core interaction stage, video sequences are separated into dynamic and static regions: dynamic regions use mask-guided sparse attention for strong interaction to capture salient object features, while static regions use a lightweight weak interaction branch to reduce redundant computation and prevent semantic interference from non-informative modalities, maintaining background stability. Experiments on three benchmark datasets show that MAVFusion outperforms current state-of-the-art methods in both single-frame metrics and temporal consistency, achieving an effective balance between inference speed and fusion quality.

\section*{Acknowledgement}

This work was supported in part by the Basic and Applied Basic Research of Guangdong Province under Grant
2023A1515140077, in part by the Natural Science Foundation of Guangdong Province under Grant 2024A1515011880, in part by the National Natural Science Foundation of China under Grant 52374166 and 62201149, in part by the Research Fund of Guangdong-Hong Kong-Macao Joint Laboratory for Intelligent MicroNano Optoelectronic Technology under Grant 2020B1212030010, and in part by the Yunnan Fundamental Research Projects under Grant 202301AV070004 and Grant 202501AS070123.

%
%
\bibliographystyle{splncs04}
\bibliography{main}

\end{document}